\documentclass{tADR2e}

\begin{document}


\title{Behavioral Learning of Dish Rinsing and Scrubbing based on Interruptive Direct Teaching Considering Assistance Rate}

\author{Shumpei Wakabayashi$^{a}$$^{\ast}$, Kento Kawaharazuka$^{a}$, Kei Okada$^{a}$, and Masayuki Inaba$^{a}$\thanks{$^\ast$Corresponding author. Email: wakabayashi@jsk.imi.i.u-tokyo.ac.jp \vspace{6pt}}\\\vspace{6pt}  $^{a}${\em{Graduate School of Information Science and Technology, The University of Tokyo, 7-3-1, Hongo, Bunkyo-ku, Tokyo, Japan}};
\\\vspace{6pt}}

\maketitle

\begin{abstract}
Robots are expected to manipulate objects in a safe and dexterous way.
For example, washing dishes is a dexterous operation that involves scrubbing the dishes with a sponge and rinsing them with water.
It is necessary to learn it safely without splashing water and without dropping the dishes.
In this study, we propose a safe and dexterous manipulation system. 
The robot learns a dynamics model of the object 
by estimating the state of the object and the robot itself, 
the control input, 
and the amount of human assistance required (assistance rate)
after the human corrects the initial trajectory of the robot's hands by interruptive direct teaching. 
By back-propagating the error between the estimated and the reference value 
using the acquired dynamics model, 
the robot can generate a control input
that approaches the reference value, 
for example, so that human assistance is not required and the dish does not move excessively.  
This allows for adaptive rinsing and scrubbing of dishes with unknown shapes and properties.
As a result, it is possible to generate safe actions that require less human assistance.

\end{abstract}


\renewcommand{\thefootnote}{}
\footnotetext{This is a preprint of an article whose final and definitive form has been published in ADVANCED ROBOTICS 2024, copyright Taylor \& Francis and Robotics Society of Japan, is available online at: https://www.tandfonline.com/doi/full/10.1080/01691864.2024.2379393.}

\section{INTRODUCTION}
Humans have designed control systems for robots 
so that they can perform dexterous object manipulation. 
An example of dexterous object manipulation is rinsing and scrubbing dishes in dishwashing. 
There are many cases where dexterous manipulations lead to unexpected results, 
for example, when washing dishes, 
water splashing or dropping dishes should be avoided. 
In addition, it is necessary to adaptively change the operation trajectory according to the shape of the tableware. 

There have been previous studies on the washing of tableware.
For instance, the operation trajectories are feed-forwardly generated by recognizing water~\cite{Okada2006}. 
The more careful operation is planned referencing the characteristics of a close object information with a proximity sensor~\cite{fujimoto2009picking}. 
In these studies, although the system is able to handle a variety of tableware, 
the washing method is designed by a human in advance as an invariable program, 
and there is no mention of the case where the operation is likely to fail. 
In addition, the system does not realize the washing operation according to the recognition of the dirt on the dishes. 
How to teach the robot to perform dexterous manipulation according to the state of the object, and how to realize recovery action, represent challenges that must be addressed.
\begin{figure}[!t]
  \centering
  \setlength\abovecaptionskip{5pt}
  \includegraphics[width=0.8\columnwidth, page=2]{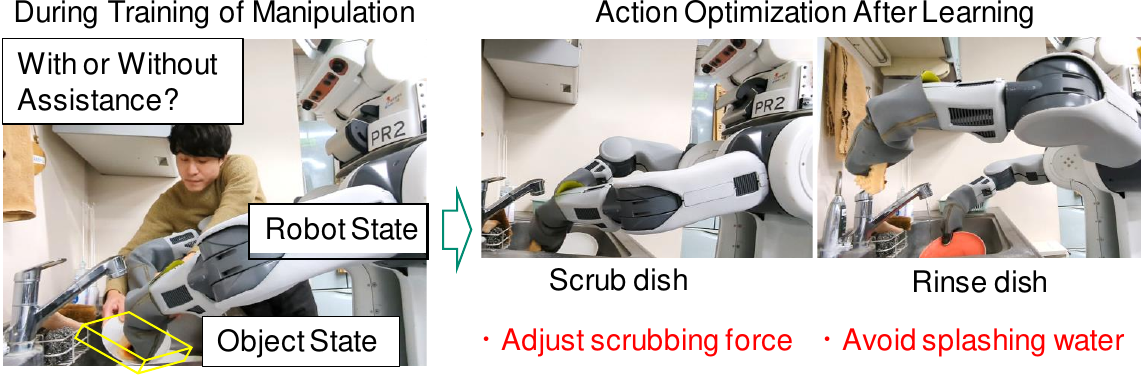}
  \caption{
      In order to manipulate an object safe and dexterous way,
      the robot executes operations based on a geometric state of the object, 
      but sometimes modified by human assistance.
      With the robot states and object states, the control input, and the amount of human assistance, the robot acquires how to manipulate it.
      After training, the robot can automatically scrub with adjusting force and rinse object avoiding splashing water.
  }
  \label{figure:overview}
\end{figure}

Dexterous and safe manipulation has been achieved by human geometric and dynamic modeling of the object to create the trajectory~\cite{dlr103272}.
Primitive actions like in-hand ball rolling~\cite{digit} and pushing, pulling and pivoting an object~\cite{Hogan2020} are modeling the object feature.
While the robot executes the actions within human expectations, 
it is difficult for the model-based planner/controller to operate adaptively for unseen objects.
In addition, it is hard to model a nonlinear controller and it is not suitable for complex action.
Therefore, research has been conducted on acquiring a planner/controller by teaching human demonstration to a robot 
such as using real-time visual recognition ~\cite{KUNIYOSHI1991}. 
It has been applied to research on complex teaching using twin arms~\cite{Caccavale2018} or satisfying geometric constraints ~\cite{Perez-DArpino2017}.
Although the teaching method previously used hidden markov model based on mixed gaussian distribution ~\cite{Tanwani2017},
deep learning has been widely used since it can deal with large dimensional data and it is easy to model the nonlinear controllers ~\cite{Osa2018, Yu2018}.
We can teach large dimensional data such as multimodal recognition that combines visual and force perception ~\cite{Noda2014,Lee2020,Anzai2020}. 
Teaching through deep learning enables the execution of complex actions such as knotting strings and wiping bowls ~\cite{Suzuki2021, Saito}. 
These techniques are often referred to as imitation learning and fall specifically under the category of behavior cloning ~\cite{pomerleau1991efficient}.

Behavior cloning has limitations.
It may not handle unknown situations or data outside the training distribution, resulting in ambiguous behavior. 
For example, in a reaching task with an obstacle in the middle of the trajectory to the goal, an agent may only learn to follow the right or left path without taking the obstacle into account. 
During inference, if the obstacle moves to the left or right, the agent may blindly follow the learned trajectory and collide with the obstacle.
%
To address this problem, Dataset Aggregation (DAgger) has been proposed ~\cite{ross2011reduction}. 
DAgger is an on-policy approach that iteratively re-trains by combining the learned policy with new data acquired during execution. 
However, DAgger can have high sample costs and may produce suboptimal performance due to redundant, low-information states. 
Furthermore, since imitation learning, even DAgger, focuses solely on mimicking the demonstrated behavior, 
it does not capture the purpose of the task. 
As a result, the reference trajectory may be difficult to interpret, 
and the agent may behave in a way that is detrimental to its objectives.
This further emphasizes the need for task-oriented learning methods.

Here, we propose an off-policy task-oriented approach that combines behavior cloning and backpropagation in a neural network.
Our behavior cloning is not trained from the start but set an initial trajectory from the geometric modeling.
We consider it easier to first roughly design the trajectory of the operation and then modify it with our temporal teaching.
In other words, we give the robot an initial trajectory based on the geometric modeling of the object and we directly move its arm to correct the trajectory.
Typical behavior cloning requires a large amount of data to learn all action sequences from a teacher.
On the other hand, we can reduce the amount of teacher data by combining an initial trajectory.
Our task-oriented approach with backpropagation in a neural network is able to adapt to new situations.
A typical strategy involves switching between model-based and learned controllers when encountering unexpected situations during object manipulation. 
For instance, a robot may predict an unforeseen state based on visual and force information ~\cite{Suzuki2021}.
In their study, the robot returns to its initial pose upon detecting danger but is expected to continue operating while evading anormal situations.
Although motion planning is feasible when actions to be taken in dangerous situations are explicitly known in the real environment~\cite{Srivastava2014a, Zimmermann2020},
sometimes it is unclear which action to take in the event of unforeseen occurrences.
In this study, we build a system that automatically and implicitly modifies behavior in response to undesirable situations.
This adaptive method is achieved by back-propagation to move closer to the target state~\cite{Tanaka2021} or moves away from the state to be avoided~\cite{Kawaharazuka2021}.
To illustrate our approach, 
let's consider the washing task, where dexterous manipulation requires reaching the arm to the dirt, rubbing it with decent force, and splashing the whole dish, 
while safe manipulation requires rubbing smoothly to avoid accidents and not disturbing the holding dish. 
The robot can achieve the reaching and holding requirements by just providing the position of the goal, while safe requirements can be expressed as the need for human assistance.
The robot is made aware that while human intervenes, the robot is attempting to output an unsafe behavior, and conversely, that while human does not intervene, the robot is executing a safe behavior.
That safe confidence can be expressed in the assistance rate as a probabilistic value.
Our contributions can be summarised as follows:
\begin{itemize}
\item
    The robot can acquire dexterous and safe manipulation by interruptive direct teaching against an initial trajectory.
\item
    The robot can accomplish the washing task by back-propagating the difference regarding the goal states and the assistance rate. 
\end{itemize}

\section{AUTOREGRESSIVE MODEL LEARNING WITH INTERRUPTIVE HUMAN ASSISTANCE AND MANIPULATION MODIFICATION DURING OPTIMIZATION} \label{sec:theory}

To achieve safe and dexterous operation, 
we propose an autoregressive dynamics model including interruptive human assistance and an optimization that modifies the control input so that human assistance is less necessary.

\subsection{Overview of an Autoregressive Model and Optimization of Control Inputs} \label{subsec:overview}
We consider learning an autoregressive model 
that approximates the function $\bm{f}$ as shown below. 
\begin{equation}
    \bm{x}_{t+1} = \bm{f}(\bm{x}_{1:t} | \bm{W_{f}})
\end{equation}
Note that $\bm{x}_{1:t}:=(\bm{x}_1, \bm{x}_2, \ldots, \bm{x}_{t})$.
$\bm{W_{f}}$ denotes the network weight of $\bm{f}$.
In this study, $\bm{x} = (\bm{s}, \bm{u}, p)$, 
where $\bm{s}$ denotes the state consisting of the robot's state $\bm{s}_{robot}$ and the object's state $\bm{s_{obj}}$.
$\bm{u}$ is the control input of the robot. 
$p$ is the amount of need for human assistance,
which takes the probabilistic value of $0 \leq p \leq 1$. 
$p = 0$ means no human assistance is needed, 
and $p = 1$ means human assistance is needed. 
The robot executes a geometrically generated manipulation trajectory relative to the object, 
and gives $p = 1$ when the human temporarily 
corrects the motion during execution, 
and $p = 0$ when the human does not.
We use the acquired data $(\bm{s}_{t}, \bm{u}_{t}, p_{t})$ 
to learn the function $\bm{f}$ that the network approximates.
After training, the control input $\bm{u}_{t}$ is optimized using the learned $\bm{f}$ as shown in the following equation.
$Loss$ is the loss function such as mean square error or binary cross entropy error between $\bm{x}_{t+1}^{ref}$ and $\bm{f}(\bm{x}_{1:t} | \bm{W_{f}})$.
\begin{equation}
    \bm{u}_{t} = argmin_{\bm{u}} Loss(\bm{x}_{t+1}^{ref}, \bm{f}(\bm{x}_{1:t} | \bm{W_{f}}))
\end{equation}

\subsection{Training an Autoregressive Model} \label{subsec:train}
\begin{figure}[!t]
  \centering
  \setlength\abovecaptionskip{5pt}
  \includegraphics[width=0.6\columnwidth, page=3]{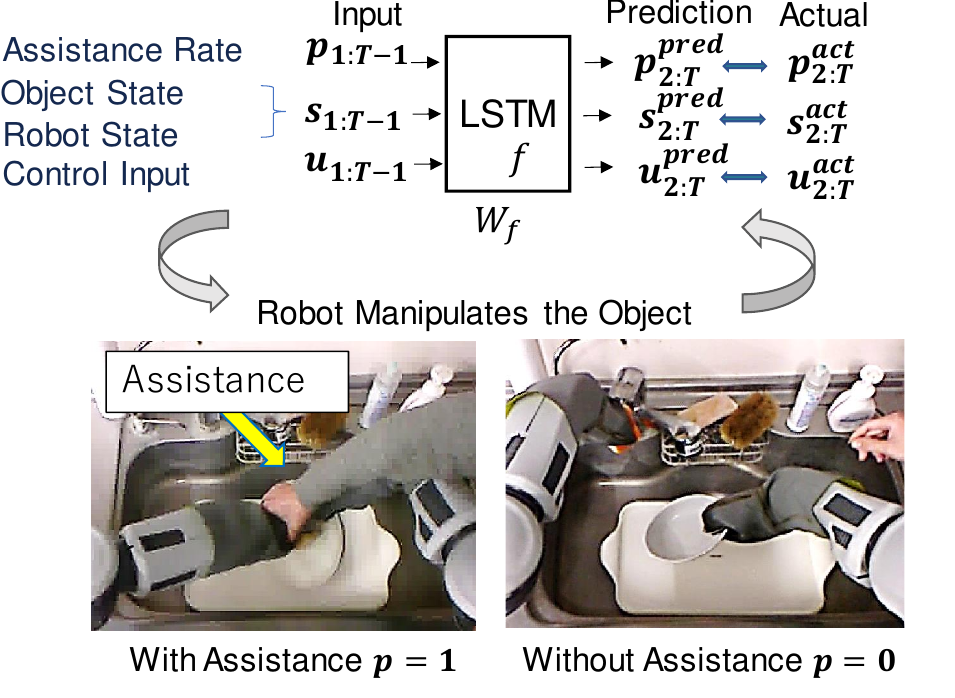}
  \caption{
      Training phase: Collecting the datasets consisted of robot states, object states and control input. Calculating the loss with predicted data and the actual data.
  }
  \label{figure:wash_train}
\end{figure}

As shown in \figref{wash_train} 
the error $L$ is calculated and the weights $\bm{W}$ of model $\bm{f}$ are learned as shown in the following equation.
\begin{align} 
    &\bm{x}_{2:T}^{pred} = \bm{f}(\bm{x}_{1:T-1} | \bm{W_{f}}) \label{xpt}\\ 
    &L_{\bm{s}} = MSE(\bm{s}_{2:T}^{pred}, \bm{s}_{2:T}^{act})\\
    &L_{\bm{u}} = MSE(\bm{u}_{2:T}^{pred}, \bm{u}_{2:T}^{act})\\
    &L_{p} = BCE(p_{2:T}^{pred}, p_{2:T}^{act})\\
    &L = \alpha_{\bm{s}} L_{\bm{s}} + \alpha_{\bm{u}} L_{\bm{u}} + \alpha_{p} L_{p} 
\end{align}
Here, $p^{actual}$ is an assistance judgment label of $0$ or $1$, where $p^{actual}=1$ means that the human is assisting and $p^{actual}=0$ means that the person is not assisting.
For example, $p = 1$ is given 
when a person corrects the trajectory of the robot's arm 
while washing dishes, 
and $p = 0$ is given otherwise. 
$T$ is the length of the manipulation sequence, 
MSE is the mean square error, 
and BCE is the Binary Cross Entropy error. 
$\bm{f}$ is trained by updating $\bm{W_{f}}$ with back propagation of the errors using Adam optimization method~\cite{Kingma2015}. 
The batch size is $C^{train}_{batch} = 4$
and the number of epochs is $C^{train}_{epoch} = 100000$.

\begin{figure}[!t]
  \centering
  \setlength\abovecaptionskip{5pt}
  \includegraphics[width=0.65\columnwidth, page=4]{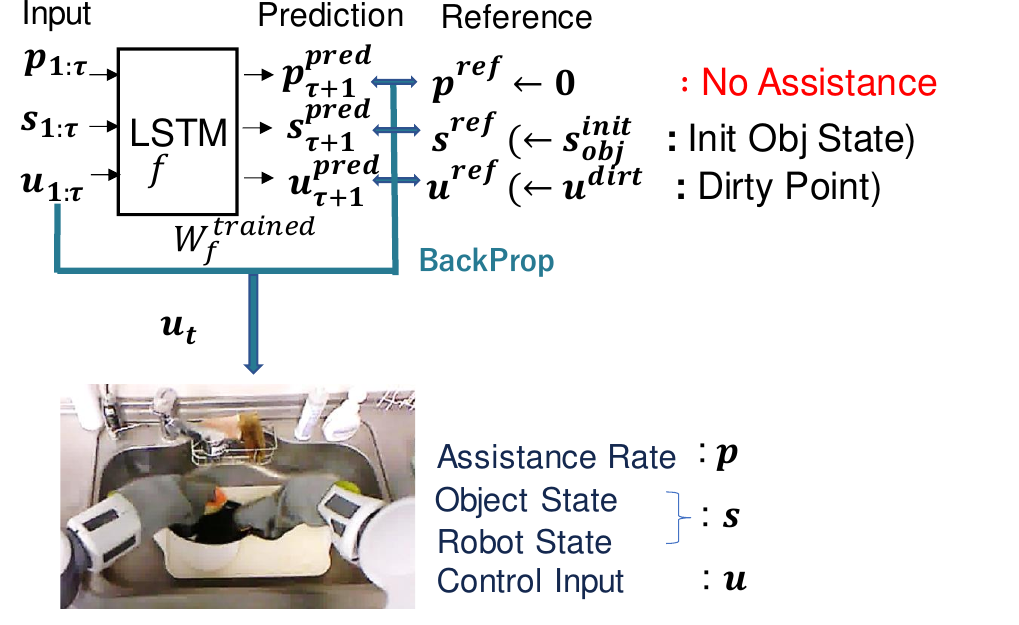}
  \caption{
      Optimization phase:  Modifying the angle vectors of the arms with reference using backpropagation of the network.
  }
  \label{figure:wash_opt}
\end{figure}

\subsection{Trajectory Optimization by Back Propagation of an Autoregressive Model} \label{subsec:opt}
The trajectory optimization is performed using the trained model $\bm{f}$ as shown in \figref{wash_opt}. 
The initial values of the run are given 
as the initial state and the joint angles $(\bm{s}, \bm{u})$, 
and the label $p = 0$, 
meaning the robot does not require human assistance.

\begin{align}
    &\bm{x}_{\tau +1}^{pred} = \bm{f}(\bm{x}_{1:\tau} | \bm{W_{f}}^{trained})\label{xpo} 
\end{align}
We calculate the error between the prediction result of the model and the reference value of each variable.
\begin{align}
    &L_{\bm{s},\tau} = MSE(\bm{s}_{\tau-\delta:\tau +1}^{pred}, \bm{s}_{const}^{ref}) \label{Ls} \\
    &L_{\bm{u},\tau} = MSE(\bm{u}_{\tau -\delta:\tau +1}^{pred}, \bm{u}_{const}^{ref}) \\
    &L_{p,\tau} = BCE(p_{\tau -\delta:\tau +1}^{pred}, p_{const}^{ref}) \\
    &L_{\tau} = \beta_{\bm{s}} L_{\bm{s},\tau} + \beta_{\bm{u}} L_{\bm{u},\tau} + \beta_{p} L_{p,\tau} \label{Lt}
\end{align}
$p_{const}^{ref}$ is set to all $0$s. 
In other words, optimization is performed so that human assistance is not required.
$\bm{s}_{const}^{ref}$ represents the state of the target object $\bm{s}_{obj}^{ref}$ and the robot itself $\bm{s}_{robot}^{ref}$. 
$\bm{u}^{ref}$ is optimized to be close to the reference value of the control input 
according to the dirty points on the dishes. 
$\delta$ is set to $3$ when $\tau \geq 4 $ and $\tau - 1$ otherwise. 
MSE is the mean square error and BCE is the Binary Cross Entropy error. 

Using $L_{\tau}$, 
the control input $\bm{u}_{\tau}$ is optimized with stochastic gradient descent method (SGD) with $C_{iter}^{test}$ iterations
shown in \textbf{Algorithm \ref{alg:alg2}.6} - \textbf{Algorithm \ref{alg:alg2}.8}.  
Finally, $\bm{u}_{t}$ is executed following \textbf{Algorithm \ref{alg:alg1}}.
$C_{epoch}^{test}$ denotes task sequence length.

\begin{figure}[!t]
  \begin{algorithm}[H]
    \caption{Action Execution}
    \label{alg:alg1}
      \begin{algorithmic}[1]
          \State  Obtain inital control input $\bm{\bm{u}_{0}}$ and $p_{0}=0$ 
          \State $Buf = [\varnothing]$
          \For{$t =1, \cdots, C_{epoch}^{test}$} 
          \State  Obtain $\bm{s}_{t}$ 
          \State $(\bm{u}_{t}, p_{t}) \leftarrow$ \textbf{Algorithm 2} $(\bm{s}_{t}, \bm{u}_{t-1}, p_{t-1}, Buf)$

          \State Execute $\bm{u}_{t}$
          \State Store $(\bm{s}_{t}, \bm{u}_{t}, p_{t})$ in $Buf$
	      \EndFor
    \end{algorithmic}
  \end{algorithm}
\end{figure}

\begin{figure}[!t]
  \begin{algorithm}[H]
    \caption{Dynamics Prediction and Optimization}
    \label{alg:alg2}
      \begin{algorithmic}[1]
      \Require 
          \Statex $\bm{s}_{t,1}$ : state of self and object at time $t$ 
          \Statex $\bm{u}_{t-1,1}$ : control input at time $t-1$
          \Statex $p_{t-1,1}$ : assistance rate at time $t-1$
          \Statex $Buf$ : data buffer
      \Ensure 
          \Statex $\bm{u}_{t-1, C_{iter}^{test}}$ : control input to be executed at time $t$
          \Statex $p_{t-1, C_{iter}^{test}}$ : assistance rate at time $t$

          \State $\bm{x}^{ref}_{const} = (\bm{s}^{ref}_{const}, \bm{u}^{ref}_{const}, p^{ref}_{const})$
          \For{$\tau =1,\cdots,C_{iter}^{test}$} 
            \State $\bm{x}_{\tau} = (\bm{s}_{t,\tau}, \bm{u}_{t-1,\tau}, p_{t-1,\tau})$ 
            \State $\bm{x}_{\tau +1} \leftarrow (\bm{x}_{\tau}, Buf)$ (Eq. \ref{xpo})
            \State $L_{\tau} \leftarrow (\bm{x}_{\tau +1}, \bm{x}^{ref}_{const})$ (Eq. \ref{Ls} - Eq. \ref{Lt})
            \State $\bm{g}_{\tau} = \partial{L_{\tau}} / \partial{\bm{u}_{t-1, \tau}}$
            \State $\bm{u}_{t-1, \tau} \leftarrow \bm{u}_{t-1, \tau} - \epsilon \bm{g}_{\tau} / \|\bm{g}_{\tau}\|$
            \State  $\qquad s.t.\, \bm{u}_{min}^{threshold}\leq \bm{u}_{t-1, \tau}\leq \bm{u}_{max}^{threshold}$
	      \EndFor
          \State \Return $\bm{u}_{t-1, C_{iter}^{test}}, p_{t-1, C_{iter}^{test}}$
    \end{algorithmic}
  \end{algorithm}
\end{figure}
\subsection{Detailis of an Autoregressive Model}
We use Long Short-Term Memory (LSTM)~\cite{Hochreiter1997} as our autoregressive model $\bm{f}$.
LSTM is capable of learning long-term dependencies.
The network input, output and hidden layer are $39$ dimensions.

\section{PATH PLANNING FOR AUTOREGRESSIVE MODEL VALIDATION} \label{section:path}
In order to confirm the usefulness of the optimization 
using the autoregressive model proposed in \subsecref{overview}, 
we conducted a preliminary experiment 
in path planning to avoid dynamic obstacles. 

\begin{figure}[!t]
  \centering
  \setlength\abovecaptionskip{5pt}
  \includegraphics[width=0.9\columnwidth, page=5]{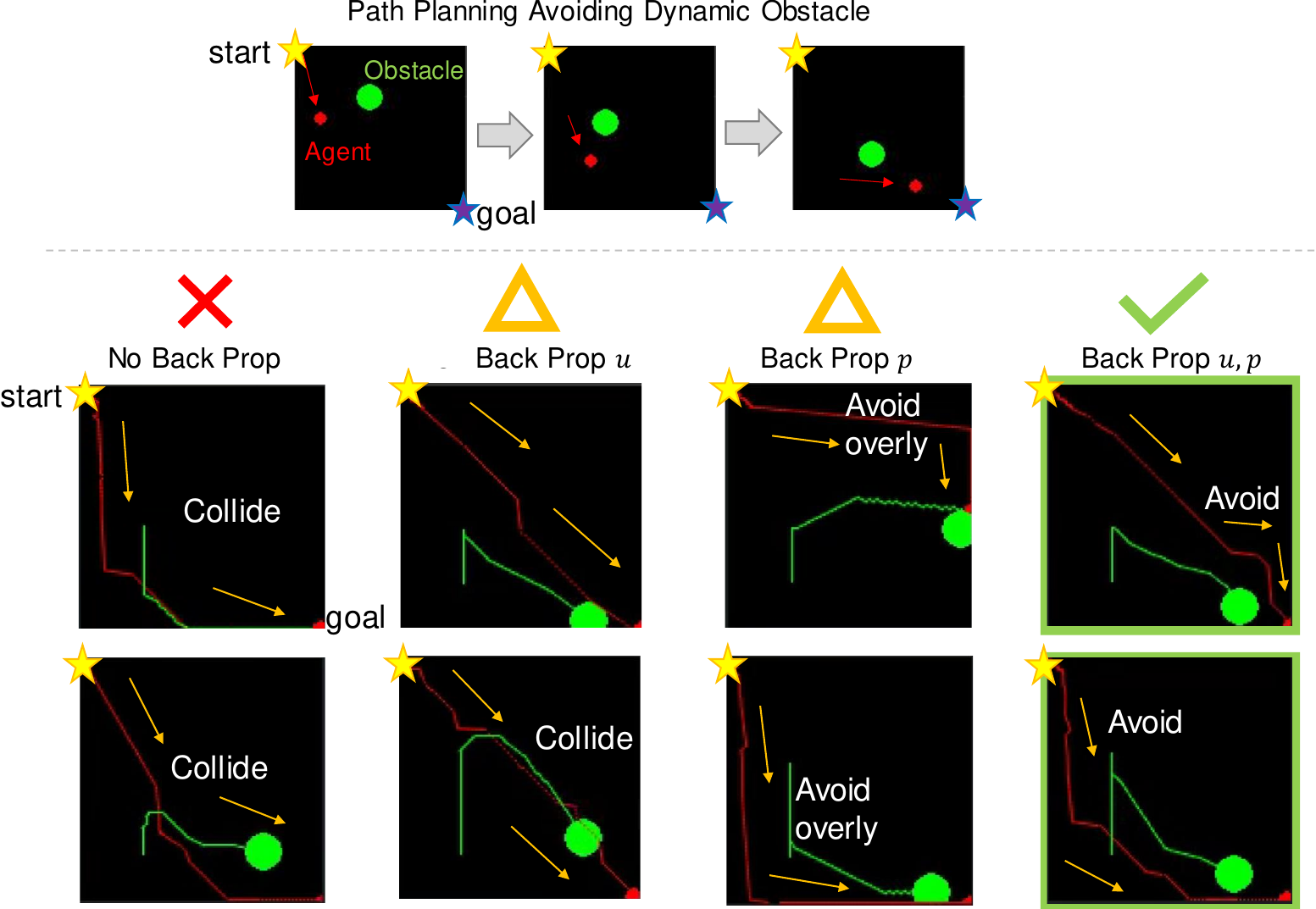}
  \caption{
      Path planning for the autoregressive model validation. A red agent aims to reach goal from top left to bottom right avoiding a dynamic green obstacle.
      The pictures show the moment when the agent reaches goal. 
      When both $u$ and $p$ are backpropageted, the agents avoid obstacle in a small circle and reaches goal.
  }
  \label{figure:toypro}
\end{figure}

\begin{table}[!t]
	\caption{
        Path planning results. When back propagating $u$ and $p$, the agent is more likely to reach goal avoiding obstacle with fewer step.
    }
	\label{table:path}
	\begin{minipage}[b]{1.0\columnwidth}
		\centering
		\begin{tabular}{|c|c|c|c|c|} \hline
                            & No BP        & BP $u$       & BP $p$    & BP $u$ and $p$ \\ \hline \hline
			Reach Goal?     &     88   \%  &   100 \%     &   76  \%  &   98  \%       \\ \hline 
			Avoid Obstacle? &     55   \%  &   62 \%      &   50  \%  &   90  \%       \\ \hline 
			Average Step    &     54       &   48         &   147      &   74           \\ \hline
		\end{tabular}
	\end{minipage}
\end{table}

\subsection{Dataset Collection and Model Training} \label{subsec:path_train}

The preliminary path planning experiment in this study is a game 
in which an agent moves to a goal while avoiding dynamic obstacles. 
If the agent follows the initial trajectory, 
it moves to the goal in a straight line from the start to the goal. 
If an obstacle approaches the agent 
while moving randomly along the way and touches the agent, 
the game fails. 
Therefore, a human can move the agent 
before it touches an obstacle. 
If the agent reaches the goal, 
the game is successful. 

In this game, $\bm{s}$ denotes the global coordinates of the agent and the obstacle.
$\bm{u}$ denotes the velocity vector of the agent.
We set $p = 1$ for cases where a human intervenes to manipulate the agent, 
and $p = 0$ for cases where there is no intervention.
A maze environment of size $(128, 128)$ is prepared, where point ($0, 0$) is the start point and point ($128, 128$) is the goal point. 
For each step, ($\bm{s}, \bm{u}, p$) are collected and are used as training data only if the game is successful. 
We collect $C_{dataset}^{train}=150$ sets of these successful games. 
The model is trained according to \subsecref{train}.

\subsection{Optimization and Results for Agent Behavior} \label{subsec:path_opt}
We optimize the trajectory of the agent using the trained model. 
In the optimization, 
we define the reference values of each variable according to \subsecref{opt}.
$\bm{u}^{ref}$ is the reference velocity vector from the current position to goal position direction.
$p^{ref}$ is set to $0$, i.e., the optimization is performed not to rely on the human assistance.
$\bm{s}^{ref}$ is not set since the coordinates of the agent and the obstacle at the next time is not known in advance.
The test sequence is set to no longer than $C_{epoch}^{test}=200$.

The results are shown in \figref{toypro}, which shows the trajectory of the agent to reach the goal.
The red circle is the agent and the green circle is the obstacle.
Note that this is the last moment of the game, 
and that even if the red and green paths overlap, 
they don't collide if they pass at different times.
In the four figures in the horizontal row, 
the random seed of the obstacle behavior is set the same.

As shown in the left side, 
Without error back-propagation 
and using $\bm{u}_{t+1}$ predicted from the autoregressive model,
the agent did not go around enough and collided with the obstacle. 
As shown in the second figure from the left in \figref{toypro}, 
when only back propagating the error between $\bm{u}_{t+1}^{pred}$ and $\bm{u}_{t+1}^{ref}$, 
the agent moved more linearly to the goal, 
but hit obstacles more easily. 
As shown in the second figure from the right in \figref{toypro}, 
when only back propagating the error for $p$, 
the agent moved around the obstacle 
but sometimes turned too far.
As shown in the right side of \figref{toypro}, 
when back propagating the error for both $\bm{u}$ and $p$, 
\\e agent avoided the obstacle while getting closer to the goal.

\tabref{path} shows the statical results of path planning.
As to whether the agent has reached the goal, when back propagating the error for $\bm{u}$, 
the agent reached the goal $100\%$ of the time, 
because it moved in a way that reduced the error between the goal direction and its own direction of movement.
When back propagating the error for $p$, the agent sometimes couldn't reach goal since it ovely turned around and exceeded the step limit.
As to whether the agent has avoided the obstacle, when back propagating the error for $\bm{u}$, 
The agent sometimes moved smoothly to the goal and did not collide with obstacles, but it often collided without much avoidance.
When back propagating the error for $p$,
the agent did indeed try to avoid the obstacle, 
but it became difficult to approach the goal 
and eventually collided with it.
As to the time step from start to goal, 
when back propagating the error for $\bm{u}$,
the agent reached the goal faster.
when back propagating the error for $p$,
the agent reached the goal slower.
From the above, when error back-propagating 
with respect to both $\bm{u}$ and $p$, 
the agent reached to the goal in shorter steps while avoiding obstacles.

\section{DISH WASHING EXPERIMENTS} \label{section:dish_exp}

\begin{table}[!t]
	\caption{Size [mm] of objects for training and optimization}
	\label{table:single}
	\footnotesize
	\begin{minipage}[b]{1.0\columnwidth}
		\centering
		\begin{tabular}{|c|c|c|c|c|} \hline
            & Object   & Depth & Width & Height \\  \hline \hline
            Training  & Round plate   &  245   &  245  & 22    \\ \cline{2-5}
                      & Polygonal plate&  205  &  205  & 25     \\ \cline{2-5}
                      & Polygonal bowl&  175  &   175  & 55    \\ \cline{2-5}
                      & Black bowl    &  150  &   150  & 52     \\ \cline{2-5}
                      & Star dish     &  150  &  145   & 42     \\ \cline{2-5}
                      & Ellipse dish  &  215  &   140  & 45     \\ \cline{2-5}
                      & Retangle dish &  210  &   110  & 30     \\ \cline{2-5}
                      & Fork          &  225  &    20  &  5     \\ \cline{2-5}
                      & Spoon         &  170  &    50  & 20     \\ \hline
            Test & Round plate   &  250  &     250 & 25     \\ \cline{2-5}
                      & Flat bowl     &  200  &     200 & 70     \\ \cline{2-5}
                      & Rice bowl     &  125  &     125 & 65     \\ \cline{2-5}
                      & Square plate  &  115   &    115  & 25     \\ \cline{2-5}
                      & Retangle plate&  115   &     237  & 25     \\ \cline{2-5}
                      & Ellipse dish &  155   &     255  & 50     \\ \cline{2-5}
                      & Fork          &  200  &     27  & 22     \\ \hline
		\end{tabular}
	\end{minipage}
\end{table}

\begin{figure}[!t]
  \centering
  \setlength\abovecaptionskip{5pt}
  \includegraphics[width=0.7\columnwidth, page=6]{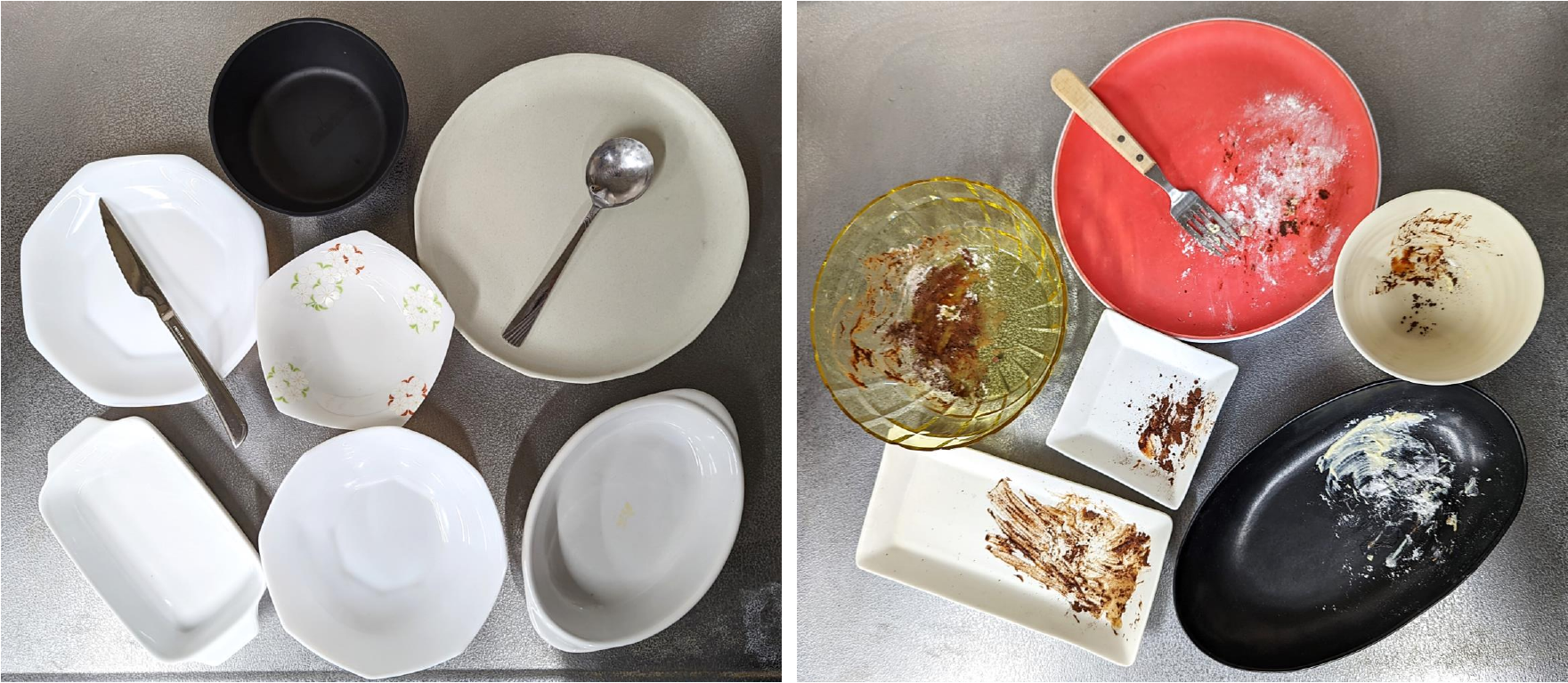}
  \caption{
	  The left picture is the dishes for washing training. The dishes are including various size and shape of plate, fork and spoon.
      The right picture is the dirty dishes for test. They are different from the dishes for training, but their domain are similar.
  }
  \label{figure:dataset}
\end{figure}

The nine types of tableware used in training are shown on the left in \figref{dataset}. 
We expect the network to generalize to a wide variety of tableware by using various shapes of tableware, a spoon and a fork.
The each size of tableware is shown in \tabref{single}.

\subsection{Collection of Datasets for Water Rinsing and Scrubbing of Tableware} \label{subsec:dish_data} 

\subsubsection{Initial Trajectory Generation Method}
First, trajectories of the end-effector relative to the object are generated. 
We obtain the bounding box of the object from the robot's recognition. 
$\bm{s}_{obj}$ is the position, orientation, and size of the bounding box, 
for a total of 10 dimensions. 
The initial trajectory of the end-effector is geometrically generated according to $\bm{s}_{obj}$. 
In the scrubbing operation, 
the right hand grasps the tableware while the left hand holds the sponge.
Here, the right hand is fixed for stability. 
A geometrical trajectory is applied only for the left holding hand and used as training data. 
A target trajectory is a sequence of randomly selected points 
on the opposite side of the upper surface of the recognized object, which is approximated by a rectangle.
In the water rinsing operation, 
the right hand grasps the dishes and the left hand operates the faucet to turn on and off the water. 
Here, the operation of the left hand is assumed to be a fixed trajectory relative to the faucet. 
Only the right hand is trained with a geometrically generated trajectory, 
which is rinsing-like heuristic motion added a random element from normal distribution.

\subsubsection{Manipulation correction with temporary human assistance}
Next, the robot continuously solves inverse kinematics for the generated initial trajectory. 
The robot executes the operation with the joint angle of the solution as the target value. 
During the execution of the robot, 
a human temporarily holds the robot's arm 
to modify the manipulation trajectory directly. 
This direct teaching is possible because the robot in this study has soft joints 
and can make flexible contact with human. 
When human judges that the initial trajectory is appropriate, 
the robot is not instructed. 
The robot continues to perform the operation 
according to the initial trajectory 
while the human is teaching it. 

In the water rinsing operation, 
the human interrupts the robot when it is about to splash water on the surface of the tableware with its right hand, 
or when it has not rinsed every corner, 
The human adjusts the angle of its right hand to adjust the surface of the plate 
that is exposed to the falling water, 
or shakes its right hand to assist the robot in spreading water to the entire tableware. 

In the scrubbing operation, 
the human judges whether the robot is scrubbing the dishes as if the sponge is pressed against the dishes 
or whether it is an unexpected operation such as almost dropping the dishes.
If the robot is not able to perform the above-mentioned actions, 
a human moves the left arm of the robot directly to assist the robot hand in holding the sponge.

\subsubsection{Data Collection and Analysis} 
The state $\bm{s}$ is composed of $24$ dimensions including $14$ dimensions of the estimated joint torque of both 7-DOF arms 
and $10$ dimensions of the object size, position and orientation. 
A object is recognized using a combination of SSD~\cite{SSD} from deep learning methods and Euclidean clustered point clouds, which is robust to occlusion.
The control input $\bm{u}$ is $14$ dimensions of the actual joint angle of both arms. 
They are acquired at least $2 Hz$.
The data are classified into the interval 
where human intervention occurs ($p = 1$) 
and the interval where no intervention occurs ($p = 0$) for the rinsing and scrubbing motions, respectively. 
Specifically, the changes in the estimated joint torques of both arms at time $t - 1$ and time $t$ are automatically labeled with a threshold value.
A series of trials were conducted with $C_{obj}^{train} = 10$ times for each of the nine tableware, 
for a total of $C_{dataset}^{train} = 90$ trials. 

\begin{figure}[!t]
  \centering
  \setlength\abovecaptionskip{5pt}
  \includegraphics[width=0.7\columnwidth, page=18]{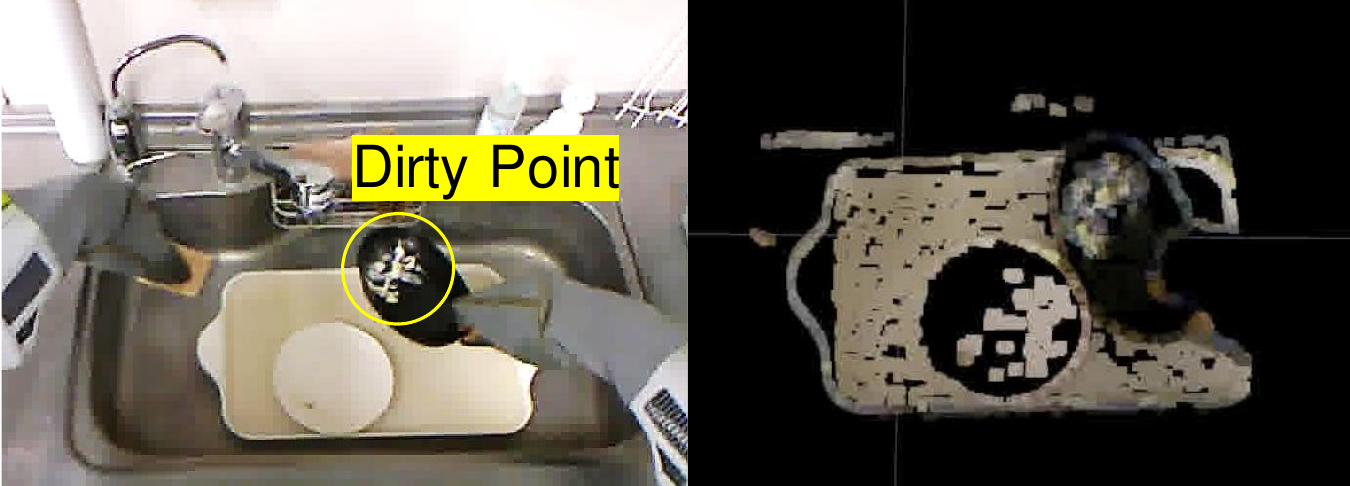}
  \caption{
      The left picture shows the point with dirt on it, which is the reference value for the trajectory.
      The right picture shows the dirt detection by image processing.
  }
  \label{figure:dirt}
\end{figure}

\subsection{Training of the Autoregressive Model for Water Rinsing and Scrubbing of Tableware} \label{subsec:wash_train} 
First, we process and augment the data.
For the control input $\bm{u}$, 
the data are augmented by adding Gaussian noise and are robust to noise.
For the estimated joint torque $\bm{s}_{robot}$, 
we smooth the torque by taking a moving average over three frames.
since there is an error due to the fact that it is an estimated value.
In addition, 
when a human assists the robot motion during training,
we want to give $\bm{s}_{robot} = 0$ 
because the estimated joint torque of a robot becomes less relevant to object manipulation. 
Therefore, the filtered value as shown in the following equation is used as the input value of $\bm{s}_{robot}$.
\begin{align}
    & \bm{s}_{robot,t}^{filtered} \leftarrow \bm{s}_{robot, t} \cdot (1 - p_{t})
\end{align}
The obtained data are sorted into scrubbing and rinsing intervals. 
The set of data in each section is learned based on the model described 
in \subsecref{train}. 
In other words, the model for the scrubbing operation 
and the model for the rinsing operation are trained respectively.

\subsection{Optimizing Manipulation in Scrubbing and Rinsing Dishes} \label{subsec:wash_opt}

The reference of $\bm{s}=(\bm{s}_{robot}, \bm{s}_{obj})$ is $\bm{s}^{ref}=(\bm{s}_{robot}^{ref}, \bm{s}_{obj}^{ref})$.
We don't set $\bm{s}_{robot}^{ref}$
because the target value of the estimated joint torque is unknown. 
$\bm{s}_{obj}^{ref}$ is set to the initial state at the time of operation, 
because the target object is not expected to move significantly by the operation.
$\bm{u}^{ref}$ is the joint angle solved by inverse kinematics at the tip of the gripper relative to the target point. 
The target point is the lower part of the faucet 
in the rinsing operation 
and the dirty point in the scrubbing operation. 
The right figure of \figref{dirt} shows the dirty point extraction.
The points with a large color difference are extracted 
by using RGB Canny Edge Detection of OpenCV. 
The distribution of the points is clustered inside the edge of the object, 
and the center of gravity of the point group in the largest cluster is used as the dirty target point. 
If $p^{predict} \geq 0.8$, 
we assume that the robot is unable to return to the original position by itself, 
and we abort the operation, place the dishes in the sink, and start over from the initial pose. 
The initial values for the inference are the data of the dish and the robot 
immediately after grasping and lifting the dishes. 
The optimization step is set to $C_{epoch}^{test}=5$.
The sequence length is set to $C_{epoch}^{test}=80$ steps for rinsing and $C_{epoch}^{test}=150$ steps for scrubbing.

\begin{figure}[!t]
  \centering
  \setlength\abovecaptionskip{5pt}
  \includegraphics[width=0.7\columnwidth, page=8]{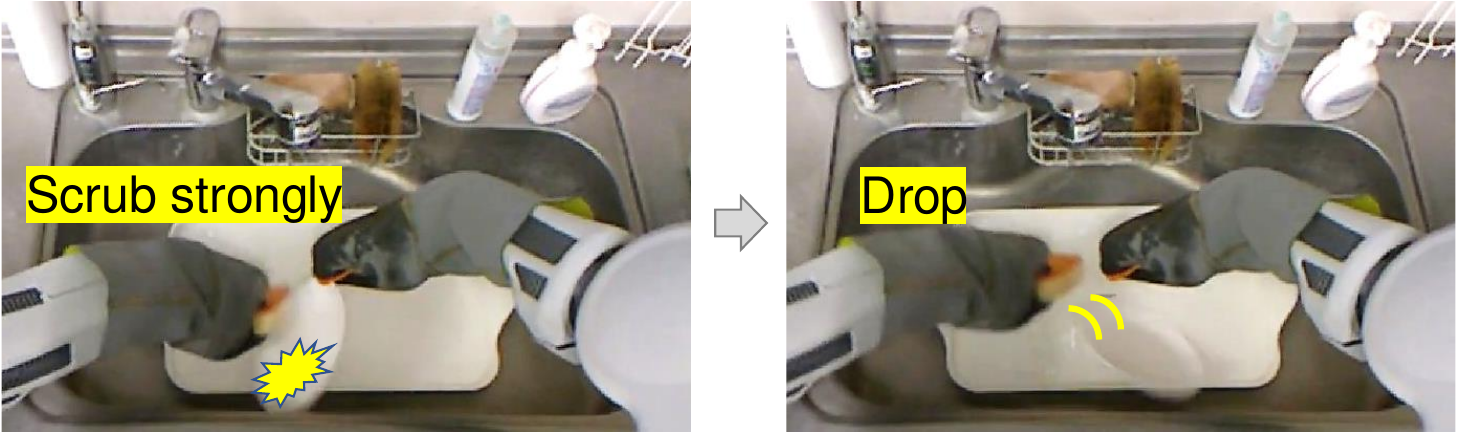}
  \caption{
      No back-propagating $p$.
      The robot executed unintended motion like scrubbing strongly.
  }
  \label{figure:bpp}
\end{figure}

\begin{figure}[!t]
  \centering
  \setlength\abovecaptionskip{5pt}
  \includegraphics[width=0.7\columnwidth, page=7]{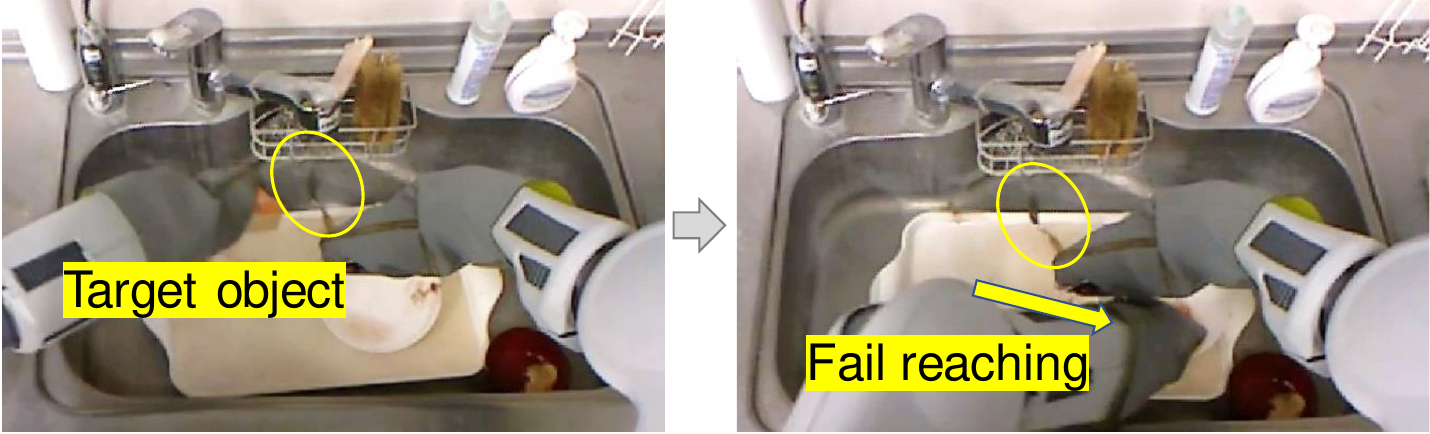}
  \caption{
      No back-propagating $\bm{u}$.
      The robot couldn't reach the target object.
  }
  \label{figure:bpu}
\end{figure}

\begin{figure}[!t]
  \centering
  \setlength\abovecaptionskip{5pt}
  \includegraphics[width=0.7\columnwidth, page=9]{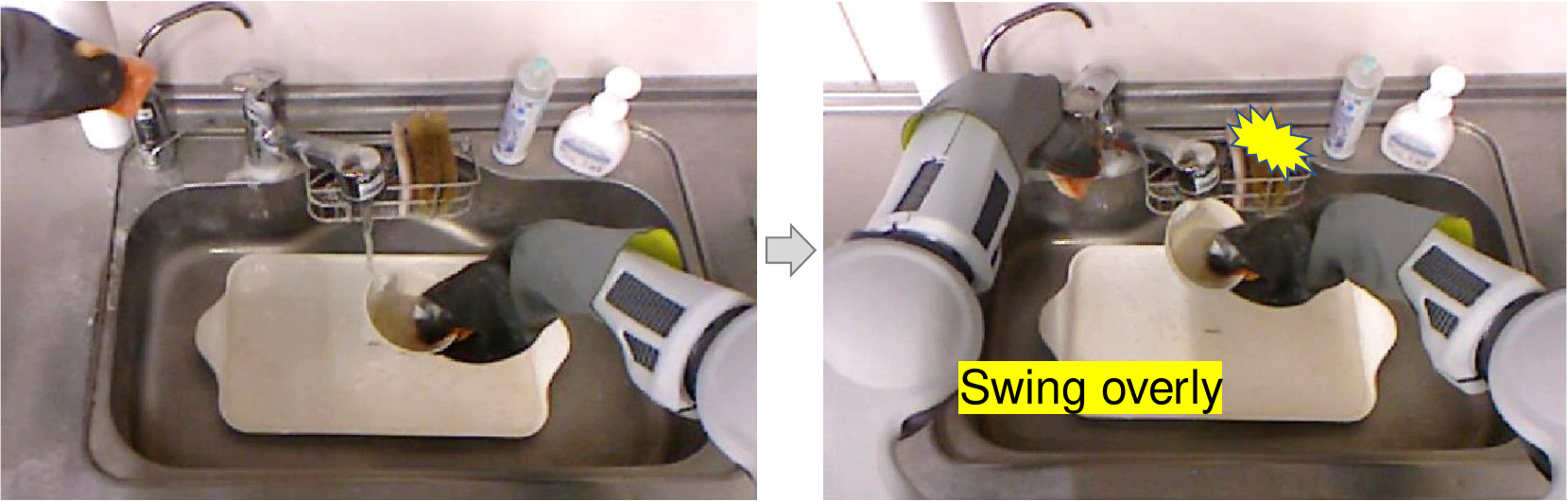}
  \caption{
      No back propagating $\bm{s}$.
      The target dish was not stable and hit the faucet when rinsing.
  }
  \label{figure:bps}
\end{figure}

\begin{figure}[!t]
  \centering
  \setlength\abovecaptionskip{5pt}
  \includegraphics[width=0.7\columnwidth, page=13]{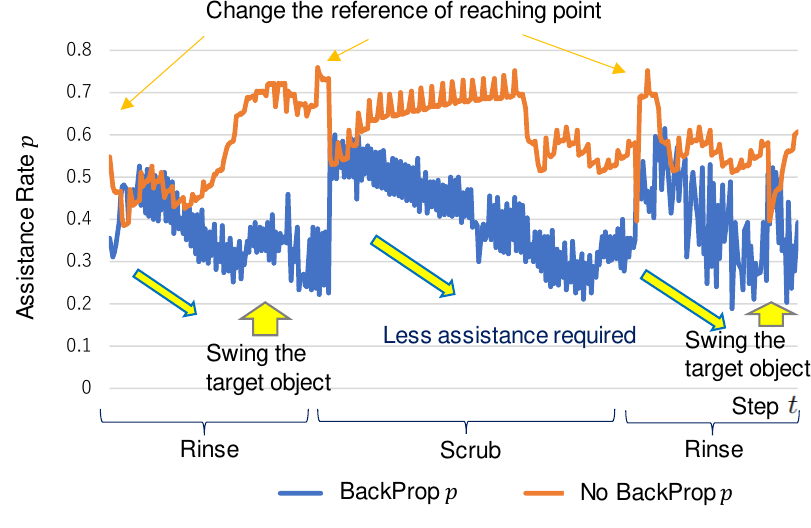}
  \caption{
	  The experimental result of the assistance rate of the manipulation.
      In the rinsing and scrubbing phases, with backpropagating $p$, the assistance rate is decreasing.
  }
  \label{figure:safety}
\end{figure}
\begin{figure}[!t]
  \centering
  \setlength\abovecaptionskip{5pt}
  \includegraphics[width=0.7\columnwidth, page=14]{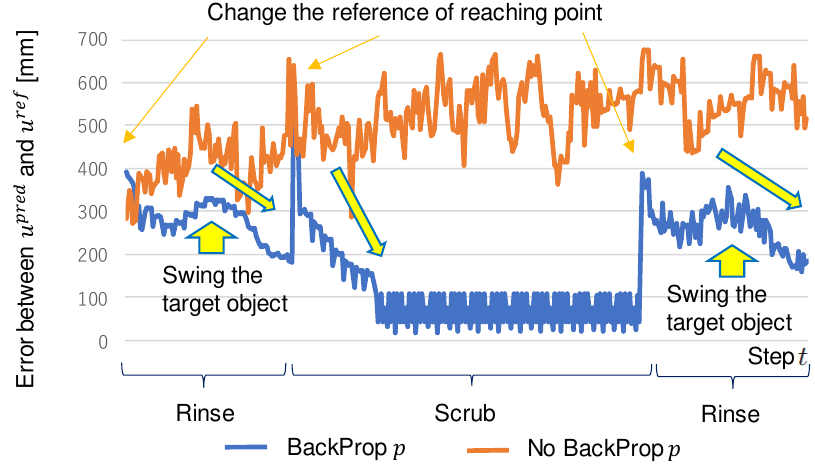}
  \caption{
      The experimental result of the error between predicted $u$ and reference $u^{ref}$.
      In the rinsing and scrubbing phases, with backpropagating $u$, the error is decreasing.
  }
  \label{figure:u}
\end{figure}
\begin{figure}[!t]
  \centering
  \setlength\abovecaptionskip{5pt}
  \includegraphics[width=0.7\columnwidth, page=15]{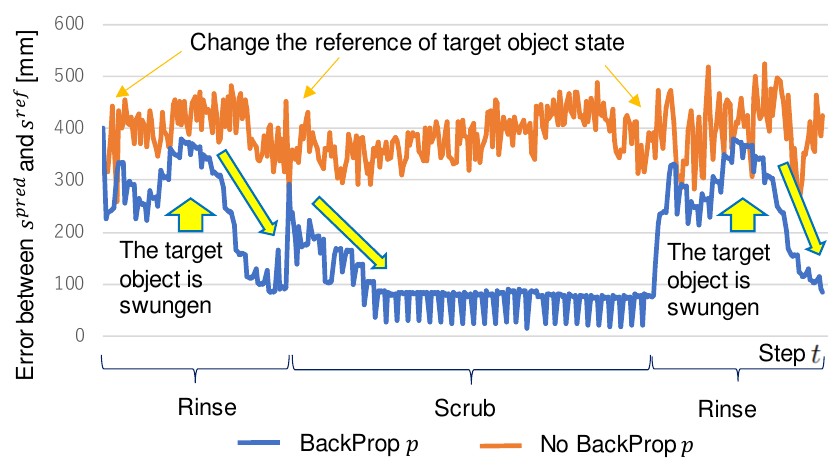}
  \caption{
      The experimental result of the error between predicted $s$ and reference $s^{ref}$.
      In the rinsing and scrubbing phases, with backpropagating $s$, the error is decreasing.
  }
  \label{figure:s}
\end{figure}

\begin{figure}[!t]
  \centering
  \includegraphics[width=1.0\columnwidth, page=16]{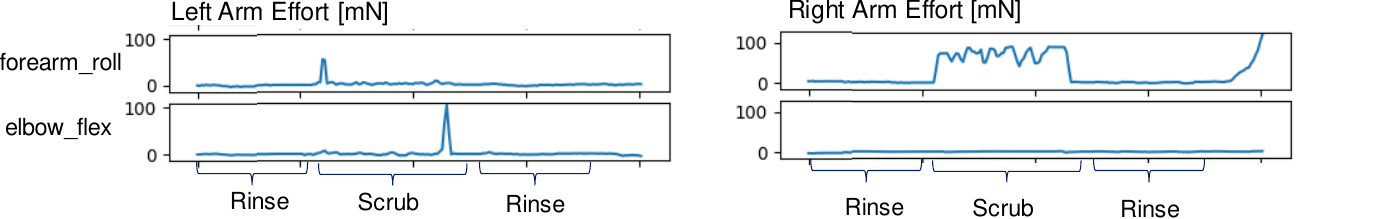}
  \caption{
      The experimental result of the joint angle effort.
      In the rinsing phase, there is no significant change in the effort,
      but in the scrubbing phase, there is a load on the right upper arm.
  }
  \label{figure:effort}
\end{figure}

\begin{figure}[!t]
  \centering
  \setlength\abovecaptionskip{5pt}
  \includegraphics[width=0.9\columnwidth, page=10]{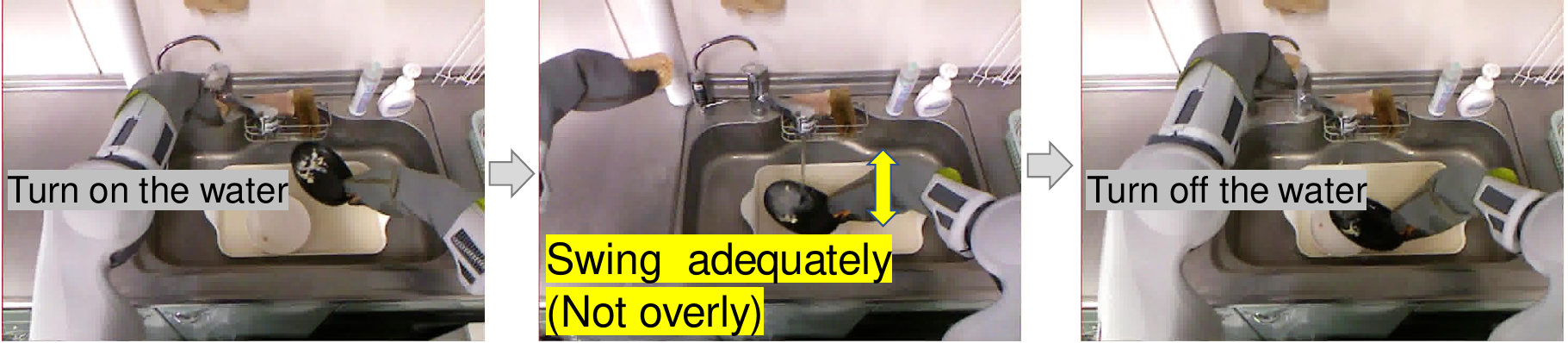}
  \caption{
      The optimization result of rinsing dish.
      The robot is rinsing while swaying its arms.
  }
  \label{figure:rinse}
\end{figure}

\begin{figure}[!t]
  \centering
  \setlength\abovecaptionskip{5pt}
  \includegraphics[width=0.9\columnwidth, page=11]{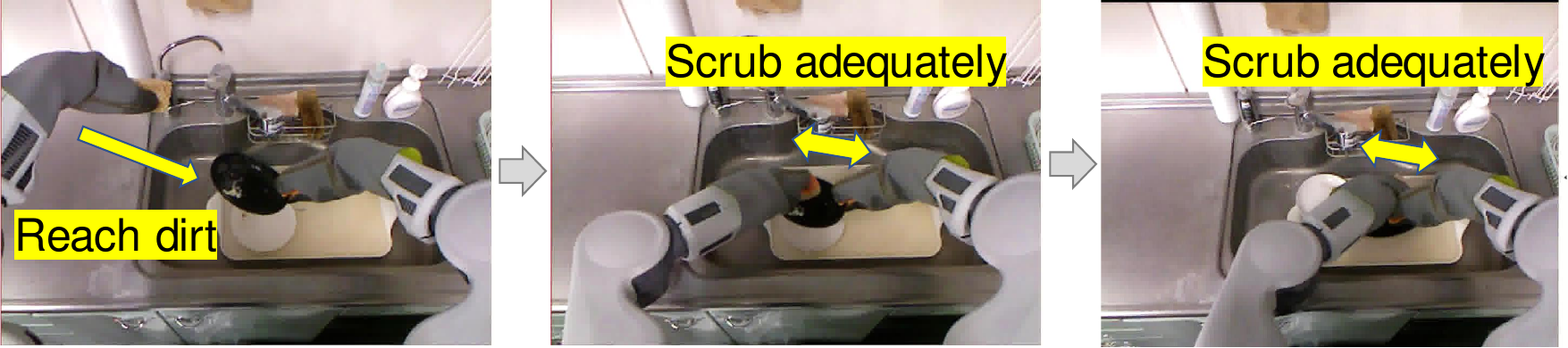}
  \caption{
      The optimization result of scrubbing dish.
      The robot scrubbed with moderate force while reaching the dirty point.
  }
  \label{figure:scrub}
\end{figure}

\begin{figure}[!t]
  \centering
  \setlength\abovecaptionskip{5pt}
  \includegraphics[width=0.9\columnwidth, page=12]{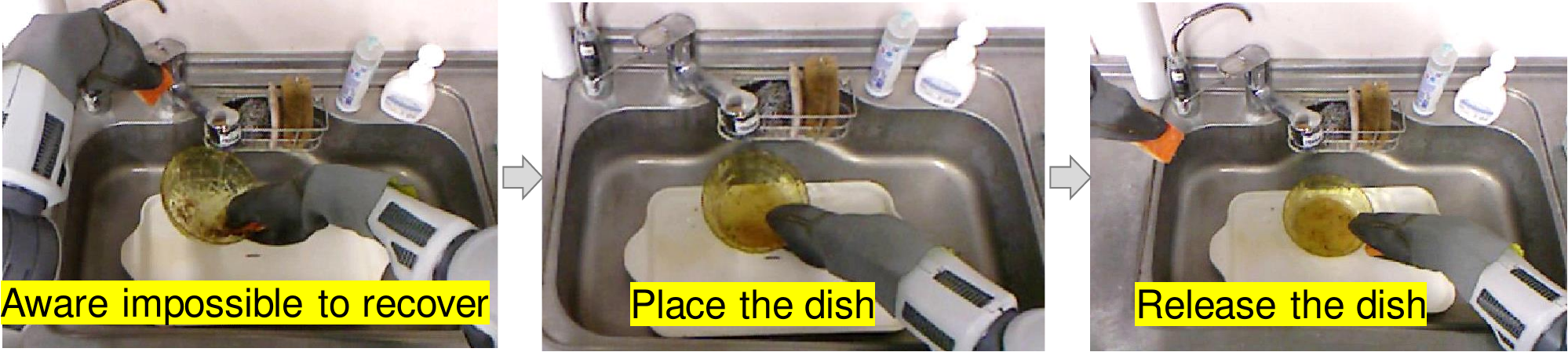}
  \caption{
      Assistance rate has exceeded the recoverable threshold, 
      so the robot placed the dish and returned to the initial pose.
  }
  \label{figure:place}
\end{figure}

\begin{figure}[!t]
  \centering
  \setlength\abovecaptionskip{5pt}
  \includegraphics[width=1.0\columnwidth, page=17]{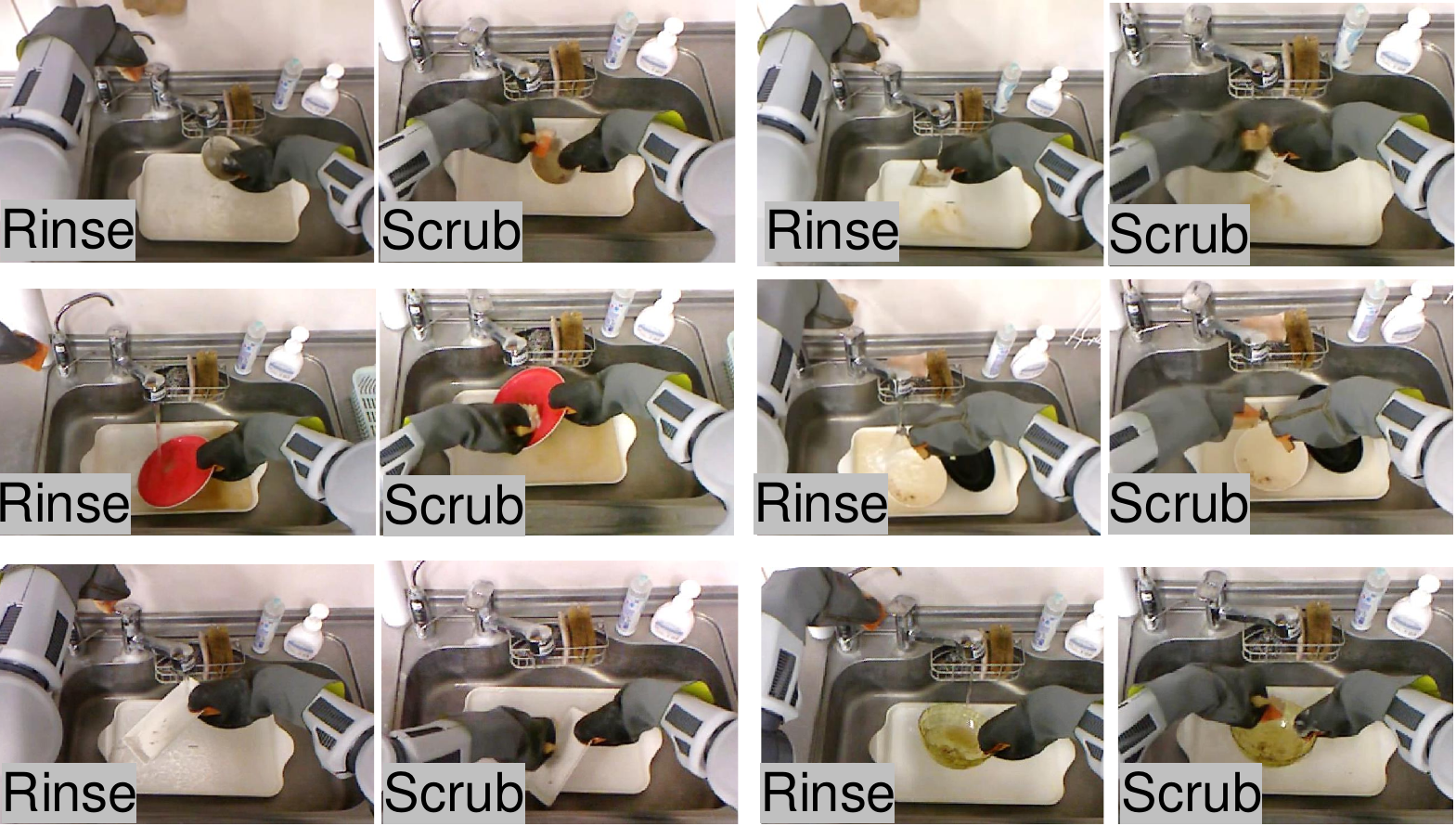}
  \caption{
      The optimization results of rinsing or scrubbing several dishes.
  }
  \label{figure:several}
\end{figure}

\subsubsection{Results}
\figref{safety} shows the results of the time series visualization of the amount of need for assistance $p$, 
during the optimization.
$p$ was high when there is no error backpropagation.
Human assistance was needed, otherwise the plate could be scraped too hard and dropped \figref{bpp}.
When there is error backpropagation, 
$p$ decreased in the rinsing and scrubbing motions 
and was optimized so that human assistance was not required.
In the rinsing phase, the hand swayed and the assistance rate increased once, 
but it quickly decreased due to error backpropagation.
\figref{u} and \figref{s} show that the errors 
between the predicted value 
and the reference values were also reduced by backpropagation.
Without back-propagating $u$, for example, the hand holding the sponge did not reach the plate \figref{bpu}.
Without back-propagating $s$, for example, the target object is shaken overly \figref{bps} 

\figref{effort} shows the estimated load of the joint angle during the operation. 
In the rinsing operation section, 
no significant changes were observed in each joint. 
In the rubbing operation section, 
the left hand was loaded on the forearm roll 
at the beginning of rubbing 
and on the elbow flex 
at the end of rubbing. 
In the right hand, 
a continuous and stable load was generated 
in the forearm roll joint during the rubbing operation. 
In the right hand, 
a continuous and stable load was generated 
at the forearm roll joint during the rubbing motion.
Note that the moderate load is not explicitly taught by a human but implicitly acquired via successful washing motion.

The sequence of the rinsing operation is shown in \figref{rinse}, 
where the tableware was moved gently to make the water drip smoothly. 
The sequence of the scrubbing operation is shown in \figref{scrub}, 
where the left hand scrubs with respect to the position of the dirt on the tableware.

There were times when the robot's motion was not the desired motion 
and the assistance rate overwhelmed the threshold, 
The robot placed tableware safely and returned to the initial pose and tried again (\figref{place}).

Finally, the robot achieved scrubbing and rinsing of all seven types of tableware, including plates, bowls, and fork (\figref{several}).

\section{DISCUSSION}
\subsection{Dexterous Manipulation}
Looking back at \subsecref{path_opt}, without back-propagating $u$, 
the agent sometimes got stuck in the place of the local optimal solution.
On the other hand, with back-propagating $u$,
the agent was more like to proceed toward the goal.

Looking back at \subsecref{wash_opt}, 
without back-propagating the error of $s$,
the robot shook the dish overly when rinsing.
On the other hand, with back-propagating it,
the robot was more likely to swing dish stably. 
Without back-propagating the error of $u$,
the robot couldn't reach a dirty point.
On the other hand, with back-propagating it,
the robot end-effector moved toward the dirt.

Considering the results of \subsecref{path_opt} and \subsecref{wash_opt},
without backpropagation, the robot executed the action associated with the trained action, 
which seemed to be a teacher action but it couldn't completely reproduce the action, 
rather it was ambiguous action mixing the several teaching actions.
On the other hand, with backpropagation, the robot modified the motion according to the reference.
This enabled the robot not only to imitate the teacher but at the same time to modify the action to complete the task.
As a results, the robot can perform dextrous manipulation such as scrubbing moderately and spreading water on the dish.

Efforts to generate more dextrous motion is a future challenge.
For example, we use PR2 robot which does not have torque sensor around the motor in the arm. 
If we can control the end effectors more precisely using a torque-controlled robot and we may be able to achieve precise movements.
In this study, however, 
a human was able to teach the robot 
while it was running because the robot had soft arms. 
Our method is unique because it is difficult to apply it in a torque-controlled robot.

\subsection{Safety Manipulation}
Looking back at \subsecref{path_opt}, without back-propagating assistance rate, the agent is more likely to collide with the obstacle.
On the other hands, with back-propagating assistance rate, the agent is more likely to avoid the obstacle.

Looking back at \subsecref{wash_opt}, 
without back-propagating assistance rate, 
the robot shook the plate and splashing the water away.
In the scrubbing phase, the robot operated unexpected movement.
On the other hands, with back-propagating the assistance rate,
the robot is more likely to tilt dish carefully when rinsing dish 
and scrub it with less human assistance.
There is a time when the robot sliped and dropped dish even when improving motion with back pagation.
In such cases, a human may be able to anticipate danger 
by observing an increase in the assistance rate.

\section{CONCLUSIONS}
In this study, we proposed a learning-based control system 
that can perform dexterous object manipulation 
without human assistance. 
We developed a system that can adaptively perform scrubbing and rinsing dirty dishes based on visual and force information. 
In the scrubbing motion, 
the trajectory and the scrubbing force should change 
according to the state of the tableware. 
In the rinsing motion, 
it is necessary to generate motion 
so that the water does not splash on the tableware 
and the water is applied to the entire tableware. 
Since the control input is unclear for these motions during the training, 
a human taught motion interruptively during the execution of the initial trajectory based on the geometric model of the object. 
During the optimization, 
the control input reference point was given 
according to the dirty area of the dish, 
and the motion was modified so that the tableware did not vibrate significantly during operation. 
At the same time, the robot memorized the situation 
in which a human assisted during teaching, 
and modified the behavior
so that no assistance was required during execution. 
When an unintended situation was judged to be irreversible, 
the operation was stopped and the dishes were placed in the sink. 
As a result, the robot scrubbed the dirty dishes with moderate force 
and rinsed them adequately, 
demonstrating more dexterous object manipulation 
that required less human assistance.

\bibliographystyle{unsrt}
\bibliography{main}

\begin{thebibliography}{10}

\bibitem{Okada2006}
Kei Okada, Mitsuharu Kojima, Yuichi Sagawa, Toshiyuki Ichino, Kenji Sato, and
  Masayuki Inaba.
\newblock {Vision based behavior verification system of humanoid robot for
  daily environment tasks}.
\newblock {\em Proceedings of the 2006 6th IEEE-RAS International Conference on
  Humanoid Robots, HUMANOIDS}, 00:7--12, 2006.

\bibitem{fujimoto2009picking}
Junya Fujimoto, Ikuo Mizuuchi, Yoshinao Sodeyama, Kunihiko Yamamoto, Naoya
  Muramatsu, Shigeki Ohta, Toshinori Hirose, Kazuo Hongo, Kei Okada, and
  Masayuki Inaba.
\newblock Picking up dishes based on active groping with multisensory robot
  hand.
\newblock In {\em RO-MAN 2009-The 18th IEEE International Symposium on Robot
  and Human Interactive Communication}, pages 220--225. IEEE, 2009.

\bibitem{dlr103272}
Daniel Leidner, Wissam Bejjani, Alin Albu-Sch{\"a}ffer, and Michael Beetz.
\newblock Robotic agents representing, reasoning, and executing wiping tasks
  for daily household chores.
\newblock In {\em International Conference on Autonomous Agents and Multiagent
  Systems (AAMAS)}, 2016.

\bibitem{digit}
Mike Lambeta, Po-Wei Chou, Stephen Tian, Brian Yang, Benjamin Maloon,
  Victoria~Rose Most, Dave Stroud, Raymond Santos, Ahmad Byagowi, Gregg
  Kammerer, Dinesh Jayaraman, and Roberto Calandra.
\newblock {DIGIT: A Novel Design for a Low-Cost Compact High-Resolution Tactile
  Sensor With Application to In-Hand Manipulation}.
\newblock {\em IEEE Robotics and Automation Letters}, 5(3):3838--3845, 2020.

\bibitem{Hogan2020}
Francois~R. Hogan, Jose Ballester, Siyuan Dong, and Alberto Rodriguez.
\newblock {Tactile Dexterity: Manipulation Primitives with Tactile Feedback}.
\newblock In {\em Proceedings - IEEE International Conference on Robotics and
  Automation}, pages 8863--8869. Institute of Electrical and Electronics
  Engineers Inc., may 2020.

\bibitem{KUNIYOSHI1991}
Yasuo Kuniyoshi, Hirochika Inoue, and Masayuki Inaba.
\newblock {Teaching by Showing: Generating Robot Command Sequences Based on
  Real Time Visual Recognition of Human Pick and Place Actions.}
\newblock {\em Journal of the Robotics Society of Japan}, 9(3):295--303, 1991.

\bibitem{Caccavale2018}
Riccardo Caccavale, Matteo Saveriano, Giuseppe~Andrea Fontanelli, Fanny
  Ficuciello, Dongheui Lee, and Alberto Finzi.
\newblock {Imitation learning and attentional supervision of dual-arm
  structured tasks}.
\newblock In {\em 7th Joint IEEE International Conference on Development and
  Learning and on Epigenetic Robotics, ICDL-EpiRob 2017}, volume 2018-Janua,
  2018.

\bibitem{Perez-DArpino2017}
Claudia Perez-D'Arpino and Julie~A. Shah.
\newblock {C-LEARN: Learning geometric constraints from demonstrations for
  multi-step manipulation in shared autonomy}.
\newblock {\em Proceedings - IEEE International Conference on Robotics and
  Automation}, (d):4058--4065, 2017.

\bibitem{Tanwani2017}
Ajay~Kumar Tanwani and Sylvain Calinon.
\newblock {A generative model for intention recognition and manipulation
  assistance in teleoperation}.
\newblock {\em IEEE International Conference on Intelligent Robots and
  Systems}, 2017-Septe:43--50, 2017.

\bibitem{Osa2018}
Takayuki Osa, Gerhard Neumann, Joni Pajarinen, J.~Andrew Bagnell, Pieter
  Abbeel, and Jan Peters.
\newblock {An algorithmic perspective on imitation learning}.
\newblock {\em arXiv}, 7(1):1--179, 2018.

\bibitem{Yu2018}
Tianhe Yu, Chelsea Finn, Annie Xie, Sudeep Dasari, Tianhao Zhang, Pieter
  Abbeel, and Sergey Levine.
\newblock {One-Shot Imitation from Observing Humans via Domain-Adaptive
  Meta-Learning}.
\newblock {\em arXiv}, 2018.

\bibitem{Noda2014}
Kuniaki Noda, Hiroaki Arie, Yuki Suga, and Tetsuya Ogata.
\newblock {Multimodal integration learning of robot behavior using deep neural
  networks}.
\newblock {\em Robotics and Autonomous Systems}, 62(6):721--736, 2014.

\bibitem{Lee2020}
Michelle~A. Lee, Yuke Zhu, Yuke Zhu, Peter Zachares, Matthew Tan, Krishnan
  Srinivasan, Silvio Savarese, Li~Fei-Fei, Animesh Garg, Animesh Garg, and
  Jeannette Bohg.
\newblock {Making Sense of Vision and Touch: Learning Multimodal
  Representations for Contact-Rich Tasks}.
\newblock {\em IEEE Transactions on Robotics}, 36(3):582--596, 2020.

\bibitem{Anzai2020}
Tomoki Anzai and Kuniyuki Takahashi.
\newblock {Deep gated multi-modal learning: In-hand object pose changes
  estimation using tactile and image data}.
\newblock {\em IEEE International Conference on Intelligent Robots and
  Systems}, pages 9361--9368, 2020.

\bibitem{Suzuki2021}
Kanata Suzuki, Momomi Kanamura, Yuki Suga, Hiroki Mori, and Tetsuya Ogata.
\newblock In-air knotting of rope using dual-arm robot based on deep learning.
\newblock In {\em 2021 IEEE/RSJ International Conference on Intelligent Robots
  and Systems (IROS)}, pages 6724--6731, 2021.

\bibitem{Saito}
Namiko Saito, Danyang Wang, Tetsuya Ogata, Hiroki Mori, and Shigeki Sugano.
\newblock Wiping 3d-objects using deep learning model based on
  image/force/joint information.
\newblock In {\em 2020 IEEE/RSJ International Conference on Intelligent Robots
  and Systems (IROS)}, pages 10152--10157, 2020.

\bibitem{pomerleau1991efficient}
Dean~A Pomerleau.
\newblock Efficient training of artificial neural networks for autonomous
  navigation.
\newblock {\em Neural computation}, 3(1):88--97, 1991.

\bibitem{ross2011reduction}
St{\'e}phane Ross, Geoffrey Gordon, and Drew Bagnell.
\newblock A reduction of imitation learning and structured prediction to
  no-regret online learning.
\newblock In {\em Proceedings of the fourteenth international conference on
  artificial intelligence and statistics}, pages 627--635. JMLR Workshop and
  Conference Proceedings, 2011.

\bibitem{Srivastava2014a}
Siddharth Srivastava, Eugene Fang, Lorenzo Riano, Rohan Chitnis, Stuart
  Russell, and Pieter Abbeel.
\newblock {Combined task and motion planning through an extensible
  planner-independent interface layer}.
\newblock {\em Proceedings - IEEE International Conference on Robotics and
  Automation}, pages 639--646, 2014.

\bibitem{Zimmermann2020}
Simon Zimmermann, Ghazal Hakimifard, Miguel Zamora, Roi Poranne, and Stelian
  Coros.
\newblock A multi-level optimization framework for simultaneous grasping and
  motion planning.
\newblock {\em IEEE Robotics and Automation Letters}, 5(2):2966--2972, 2020.

\bibitem{Tanaka2021}
Daisuke Tanaka, Solvi Arnold, and Kimitoshi Yamazaki.
\newblock Disruption-resistant deformable object manipulation on basis of
  online shape estimation and prediction-driven trajectory correction.
\newblock {\em IEEE Robotics and Automation Letters}, 6(2):3809--3816, 2021.

\bibitem{Kawaharazuka2021}
Kento Kawaharazuka, Naoki Hiraoka, Yuya Koga, Manabu Nishiura, Yusuke Omura,
  Yuki Asano, Kei Okada, Koji Kawasaki, and Masayuki Inaba.
\newblock Online learning of danger avoidance for complex structures of
  musculoskeletal humanoids and its applications.
\newblock In {\em 2020 IEEE-RAS 20th International Conference on Humanoid
  Robots (Humanoids)}, pages 349--355, 2021.

\bibitem{Kingma2015}
Diederik~P. Kingma and Jimmy~Lei Ba.
\newblock {Adam: A method for stochastic optimization}.
\newblock {\em 3rd International Conference on Learning Representations, ICLR
  2015 - Conference Track Proceedings}, pages 1--15, 2015.

\bibitem{Hochreiter1997}
Sepp Hochreiter and J\"{u}rgen Schmidhuber.
\newblock {Long Short-Term Memory}.
\newblock {\em Neural Computation}, 9(8):1735--1780, 11 1997.

\bibitem{SSD}
Wei Liu, Dragomir Anguelov, Dumitru Erhan, Christian Szegedy, Scott Reed,
  Cheng-Yang Fu, and Alexander~C Berg.
\newblock {SSD: Single Shot MultiBox Detector}.
\newblock In Bastian Leibe, Jiri Matas, Nicu Sebe, and Max Welling, editors,
  {\em Computer Vision -- ECCV 2016}, pages 21--37, Cham, 2016. Springer
  International Publishing.

\end{thebibliography}

\end{document}